\documentclass[sigconf]{acmart}

\usepackage[utf8]{inputenc} 
\usepackage[T1]{fontenc}    
\usepackage{hyperref}
\makeatletter
\renewcommand{\sectionautorefname}{\S\,\@gobble}
\renewcommand{\subsectionautorefname}{\S\,\@gobble}
\renewcommand{\subsubsectionautorefname}{\S\,\@gobble}
\makeatother      
\usepackage{booktabs}       
\usepackage{amsfonts}       
\usepackage{nicefrac}       
\usepackage{microtype} 
\usepackage{xcolor,colortbl}         
\usepackage{amsthm}
\usepackage{amsmath}
\usepackage{breqn}
\usepackage{adjustbox}
\usepackage{float}
\usepackage{graphicx}
\usepackage{subcaption}
\usepackage{tablefootnote}
\usepackage{xspace}
\usepackage{bm}
\usepackage{pifont}
\usepackage{balance}

\usepackage[super]{nth}
\newcommand{\cmark}{\textcolor{green}{\text{\ding{51}}}}
\newcommand{\xmark}{\textcolor{red}{\text{\ding{55}}}}

\usepackage[draft,inline,nomargin,index]{fixme}
\fxsetup{theme=colorsig,mode=multiuser,inlineface=\itshape,envface=\itshape}
\FXRegisterAuthor{aj}{aaj}{Avijit}
\FXRegisterAuthor{cw}{acw}{Christo}

\newcommand{\eg}{e.g.,\ }

\newcommand{\ie}{i.e.,\ }

\newcommand{\aka}{a.k.a.\ }

\newcommand{\fulltitle}{FairCanary: Rapid Continuous Explainable Fairness}
\newcommand{\fullkeywords}{fairness, drift, model explanation, continuous measurement}

\definecolor{linkColor}{RGB}{6,125,233}
\hypersetup{%
  pdftitle={\fulltitle},
  pdfauthor={},
  pdfkeywords={\fullkeywords},
  bookmarksnumbered,
  pdfstartview={FitH},
  colorlinks,
  citecolor=black,
  filecolor=black,
  linkcolor=black,
  urlcolor=linkColor,
  breaklinks=true,
  hypertexnames=false
}

\copyrightyear{2022}
\acmYear{2022}
\setcopyright{acmlicensed}\acmConference[AIES'22]{Proceedings of the 2022 AAAI/ACM Conference on AI, Ethics, and Society}{August 1--3, 2022}{Oxford, United Kingdom}
\acmBooktitle{Proceedings of the 2022 AAAI/ACM Conference on AI, Ethics, and Society (AIES'22), August 1--3, 2022, Oxford, United Kingdom}
\acmPrice{15.00}
\acmDOI{10.1145/3514094.3534157}
\acmISBN{978-1-4503-9247-1/22/08}

\begin{document}

\title{\fulltitle}

\author{Avijit Ghosh}
\email{ghosh.a@northeastern.edu}
\orcid{0000-0002-8540-3698}
\affiliation{%
  \institution{Northeastern University}
  \city{Boston}
  \state{MA}
  \country{USA}
}
\authornote{Authors contributed equally to this work. Authors were employed at Fiddler Labs Inc. when the majority of this work was conducted.}

\author{Aalok Shanbhag}
\email{aalokshanbhag@gmail.com}
\orcid{0000-0003-2765-4816}
\affiliation{%
  \institution{Snap Inc.}
  \city{Mountain View}
  \state{CA}
  \country{USA}
}
\authornotemark[1]

\author{Christo Wilson}
\email{cbw@ccs.neu.edu}
\orcid{0000-0002-5268-004X}
\affiliation{%
  \institution{Northeastern University}
  \city{Boston}
  \state{MA}
  \country{USA}
}

\renewcommand{\shortauthors}{Ghosh, et al.}

\begin{abstract}
  Systems that offer \textit{continuous model monitoring} have emerged in response to (1) well-documented failures of deployed Machine Learning (ML) and Artificial Intelligence (AI) models and (2) new regulatory requirements impacting these models. Existing monitoring systems continuously track the performance of deployed ML models and compute feature importance (\aka \textit{explanations}) for each prediction to help developers identify the root causes of emergent model performance problems.
  

  We present Quantile Demographic Drift (QDD), a novel model bias quantification metric that uses quantile binning to measure differences in the overall prediction distributions over subgroups. QDD is ideal for continuous monitoring scenarios, does not suffer from the statistical limitations of conventional threshold-based bias metrics, and does not require outcome labels (which may not be available at runtime). We incorporate QDD into a continuous model monitoring system, called FairCanary, that reuses existing explanations computed for each individual prediction to quickly compute explanations for the QDD bias metrics. This optimization makes FairCanary an order of magnitude faster than previous work that has tried to generate feature-level bias explanations. 
\end{abstract}

\ccsdesc[500]{Information systems~Data analytics}
\ccsdesc[500]{Computing methodologies~Machine learning}


\keywords{\fullkeywords}

\maketitle

\section{Introduction}


\begin{table*}[t]
    \centering
   \begin{tabular}{llcc}
\toprule
\textbf{Metric/Framework}      & \textbf{Related Terms}  & \textbf{CO?} & \textbf{E?}  \\ \midrule
Demographic parity~\cite{dwork2012fairness}             & mean difference, demographic parity, disparate treatment                     & \xmark                          & \xmark                     \\ 
Conditional statistical parity~\cite{corbett2017algorithmic} & statistical parity, conditional procedure accuracy, disparate treatment                           & \xmark                          & \xmark                     \\ 
Equalized odds~\cite{hardt2016equality}                & equalized odds, false positive/negative parity, disparate treatment                                   & \xmark                          & \xmark                     \\ 
Equal opportunity~\cite{hardt2016equality}              & equality of opportunity, individual fairness, disparate treatment                                      & \xmark                          & \xmark                     \\ 
Counterfactual fairness~\cite{kusner2017counterfactual,black2020fliptest}       & counterfactual fairness, disparate treatment, fliptest                        & \xmark                          & \xmark                     \\ 
Statistical independence~\cite{grari2019fairness}       & HGR coefficient, independence                                                                            & \cmark                         & \xmark                     \\ 
Distributional difference~\cite{miroshnikov2020wasserstein}      & KL divergence, JS Divergence, Wasserstein distance                                                    & \cmark                         & \cmark                    \\ \bottomrule
\end{tabular}
    \vspace{1em}
    \caption{Summary showing whether conventional classes of fairness metrics support Continuous Output (CO) and feature-level Explanations (E). Metric families are inspired by~\citet{mehrabi2019survey} and the related terminology is from~\citet{dasfairness}.}
    \label{tab:prosconstable}
\end{table*}

Machine Learning (ML) and Artificial Intelligence (AI) models that are deployed into the field cannot guarantee consistent performance over time~\cite{schelter2018challenges}. One of the reasons for this might be that the underlying data has changed stochastically. This phenomenon, called \textit{drift}, has been well-studied in the literature, from sudden~\cite{sudden} to gradual drifts~\cite{stanley2003learning}. Drifts may also be caused by true shifts in the relationship between the underlying variables (\eg due to changes in the population over time), sampling issues~\cite{salganicoff1997tolerating}, or even bugs that impact downstream data collection.

In scenarios where a deployed ML model is making sensitive decisions, we argue that analyzing the impact of drift on the \textit{fairness} of the model is equally, if not more, important than assessing the impact of drift on traditional performance metrics like accuracy and recall. Regulators are also concerned about this issue: for example, the European Commission's recently proposed Artificial Intelligence Act states \textit{``[ML] providers should be able to process\dots special categories of personal data, as a matter of substantial public interest, in order to ensure the bias monitoring, detection and correction in relation to high-risk AI systems''} as part of a \textit{``robust post-market monitoring system.''}~\cite{eureg} Similar regulations have been proposed in New Zealand~\cite{nzreg}, Canada~\cite{canadareg}, the US~\cite{usreg}, and the UK~\cite{ukreg}.

The recognition that drift can negatively impact model performance, coupled with looming regulations, has spurred the creation of many commercial systems that offer \textit{continuous model monitoring}~\cite{neptune}. In general, these systems track live model predictions over time, alert the operator if performance metrics change substantively, and compute feature importance (\aka \textit{explanations}) for each prediction using methods like LIME~\cite{ribeiro2016should} or the Shapley Value~\cite{datta2016algorithmic,merrick2020explanation,shapnips}.\footnote{At a high-level, LIME and SHAP are \textit{surrogate models} that are trained alongside a target model and predict how changes to feature values will impact the target model's output. High-impact features are likely to be very important to the target model.} Some of these monitoring systems incorporate fairness metrics in addition to traditional performance metrics~\cite{nigenda2021amazon}.

In this paper, we present a novel model bias quantification metric called Quantile Demographic Disparity (QDD) that uses quantile binning to measure differences in the overall prediction distributions over subgroups. Because QDD is measured over continuous distributions it does not require developers to choose specific (and often ad hoc) thresholds for measuring fairness, unlike most conventional fairness metrics (see \autoref{tab:prosconstable}).\footnote{We provide evidence for why ad hoc bias detection thresholds may miss unfairness in \autoref{sec:conventional} and \autoref{sec:casestudy}.} Additionally, QDD does not require outcome labels, which may not be available at runtime. We incorporate QDD into FairCanary, a continuous model monitoring system that offers significant advantages versus state-of-the-art commercial systems that help ensure model fairness over time. In particular, FairCanary reuses explanations computed for each individual prediction to quickly compute explanations for the QDD bias metrics. This optimization makes FairCanary an order of magnitude faster than previous work that has tried to generate feature-level bias explanations~\cite{miroshnikov2020wasserstein}.

The rest of the paper is structured as follows. In \autoref{sec:background} we present an overview of \textit{concept drift} in ML, existing work on continuous fairness monitoring, and ML fairness approaches.
Next, in \autoref{sec:overview}, we introduce our system, FairCanary, present an overview of its operation and capabilities, formally define our QDD metric, and discuss how to obtain explanations for it by reusing existing prediction attributions. In \autoref{sec:casestudy} we present a synthetic case study that highlights FairCanary's capabilities, and conclude in \autoref{sec:discussion}.

\section{Background}
\label{sec:background}

We begin by reviewing the concept of drift and the problems it causes for ML models. We highlight how continuous model monitoring can be used to identify and mitigate the problems caused by drift, but also shortcomings of existing monitoring systems. Finally, we present an overview of ML fairness metrics from the literature and discuss how their shortcomings motivated us to develop novel metrics for this project.




\subsection{Drift in Machine Learning} 

Even though a ML model may pass quality control in terms of performance during training, once deployed on live data the model may encounter issues over time that degrade or destroy its performance~\cite{sculley2015hidden}. One of the main issues that can arise in deployment is \textit{drift}, which is caused by divergence between the data and context under which the model was trained, and the real-world context into which it is deployed. \textit{Data drift} occurs when the runtime data is significantly different from the training data, by virtue of the constant changing of real world data~\cite{breck2019data}. \textit{Concept drift}, in contrast, occurs when the relationship between the model output and the feature variables change~\cite{sudden,stanley2003learning,salganicoff1997tolerating}.

Scholars have noted that model performance issues caused by drift extend to questions of algorithmic fairness~\cite{barocas2021designing}, \ie the removal of unfair and unjustified biases from ML and AI systems. For example, a temporal analysis by~\citet{liu2018delayed} showed how the changing of fairness metrics over time, due to data drift, concept drift, or otherwise, could actually harm sensitive groups.

The most popular methods for detecting concept drift~\cite{bifet2007learning,gama2013evaluating} assume that the labels for the predicted variable are immediately available. This may not be feasible in practice, however, especially if the labels correspond to sensitive features of human beings. Furthermore, even if labels are immediately available, concept drift may have rendered them unreliable, thus defeating the purpose of using them to detect concept drift. Given these issues, prior work~\cite{pinto2019automatic,DBLP:journals/corr/abs-1902-02808, dos2016fast,vzliobaite2010change} has measured the drift of prediction distributions as a proxy for concept drift. 

FairCanary also uses prediction distribution drift to measure temporal unfairness. Instead of measuring the drift of the production prediction distribution against training prediction distribution, like in ~\cite{pinto2019automatic,DBLP:journals/corr/abs-1902-02808, dos2016fast,vzliobaite2010change}, we measure the shift in the prediction distributions between different protected groups. If the prediction distributions for two protected groups start diverging over time, that is an indication of unfairness.

\begin{figure*}[t]
    \begin{subfigure}[t]{0.45\textwidth}
        \centering
        \includegraphics[width=\columnwidth,keepaspectratio]{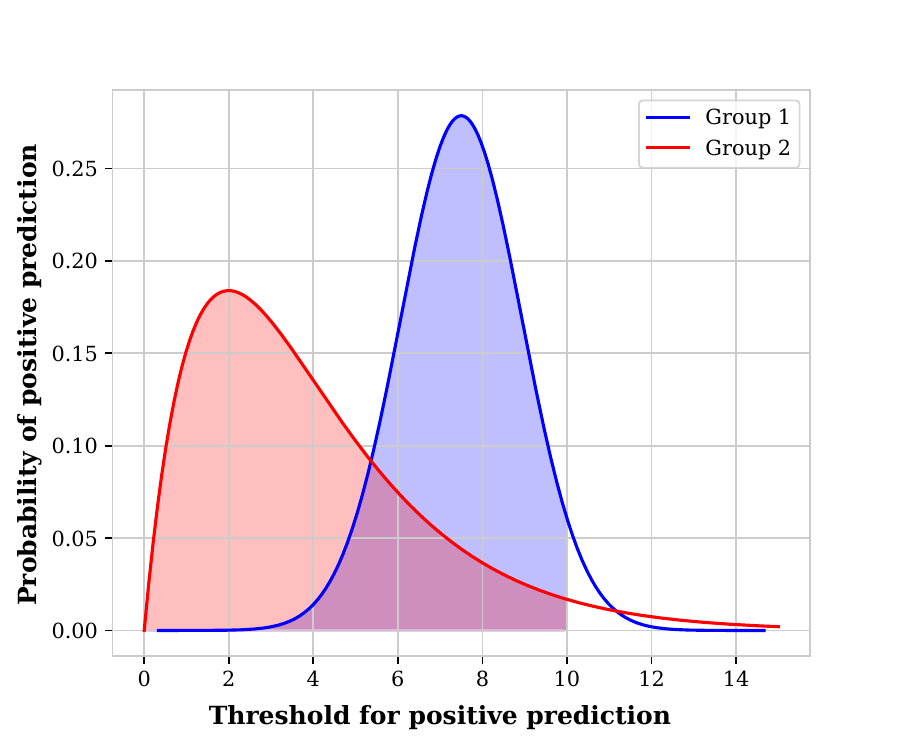}
        \caption{Probability Density Plot}
    \end{subfigure}
    \begin{subfigure}[t]{0.45\textwidth}
        \centering
        \includegraphics[width=\columnwidth,keepaspectratio]{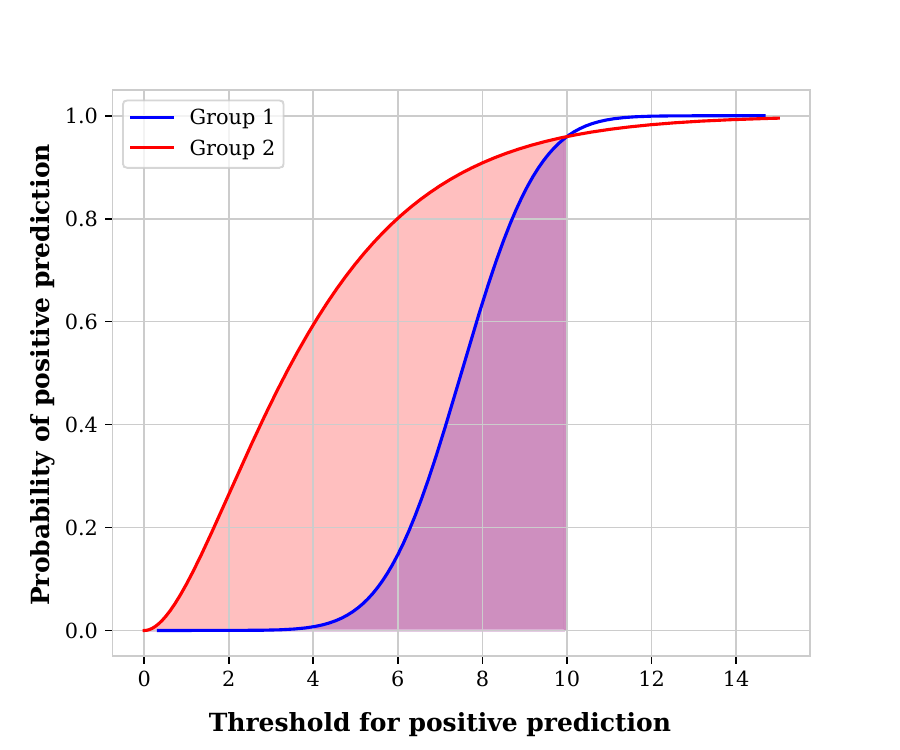}
        \caption{Cumulative Distribution Plot}
    \end{subfigure}
    \caption{Probability distribution plots for two hypothetical demographic groups. As demonstrated by the CDF plot on the right, at a threshold of \pmb{$x = 10$} the positive prediction probability for both groups is about 0.95, thereby satisfying Demographic Parity [\pmb{$P(Y^*) | D_1 = P(Y^*) | D_2$}], but this is misleading: the Wasserstein distance is nonzero since the two distributions have markedly different shapes. In contrast, moving the threshold to \pmb{$x = 8$} immediately disadvantages one group, since the positive prediction probability for group 1 falls to 0.6 while for group 2 it only falls to 0.9, thereby violating Demographic Parity.}
    \label{fig:threshold}
\end{figure*}


The primary mitigation against drift is retraining models on fresher data. Retraining may be expensive, however, so determining when to retrain models is crucial: retraining too frequently wastes (potentially substantial) resources~\cite{bender-2021-facct}, while waiting too long runs the risk of performance degradation.


\subsection{Model Monitoring and Explanations}

\textit{Continuous model monitoring} systems are designed to help developers ensure that deployed models perform as expected over time in the face of problems like drift. A number of commercial tools are available that offer model monitoring~\cite{neptune}. In general, these systems offer the following features:
\begin{itemize}
    \item Continuously record model inputs and model predictions.
    \item Measure and report traditional performance metrics over time, like precision, recall, and accuracy. Some systems also measure bias/fairness metrics.
    \item Calculate and record feature-level explanations using techniques like LIME~\cite{ribeiro2016should} or SHAP~\cite{datta2016algorithmic,merrick2020explanation,shapnips}, which are useful for post-mortem analysis if problems are observed.
    \item Generate alarms if particular metrics fall below an operator-specified threshold.
\end{itemize}
Continuous model monitoring systems are useful for uncovering a variety of issues with models at deployment time, including issues caused by drift. Once the developer has identified an issue they can apply mitigations, such as model retraining.

While virtually all of the commercially available monitoring systems explain predictions in terms of constituent features, none of them (to the best of our knowledge) offer explanations for measures of unfairness. We argue that it is equally important to understand which particular input features are responsible for causing unfairness to the model over time, especially given the ``right to explanation'' that is increasingly being enshrined in regulation~\cite{selbst2018meaningful}.

Unfortunately, the interpretation of fairness metrics in terms of the input features to the model has not been studied extensively so far. Explaining conventional fairness metrics (see \autoref{tab:prosconstable}) that rely on ground truth labels using Shapley values is possible by making the assumption that the perturbed values retain the original output label. This approach can be misleading, however, because the perturbations change the nature of the instance, and can even create Out-of-Distribution (OOD) points~\cite{kumar2020problems}. Another approach, proposed by~\citet{shanbhag2021unified}, explains differentiable distance metrics using integrated gradients, but this technique only applies to differentiable models, which limits its practical applications. Finally,~\citet{miroshnikov2020wasserstein} developed methods to explain the Wasserstein-1 distance using a Shapley value formulation. However, this approach also suffers from practical challenges: (1) it requires that explanations be computed for every possible pair of protected groups, and (2) it is computationally challenging to compute Shapley values over large samples. 

FairCanary is closely related to the work by~\citet{miroshnikov2020wasserstein}, with a couple of key differences. While~\citet{miroshnikov2020wasserstein} calculated fairness explanations from scratch using Shapley-based methods, for FairCanary we assume that a system that continuously generates prediction explanations, like the systems in~\cite{neptune}, are already available. FairCanary sits on top of such a system and reuses these existing prediction explanations to generate fairness explanations in linear time (\autoref{sec:attribution}).



\begin{figure*}[t]
    \centering
    \includegraphics[width=\textwidth,keepaspectratio]{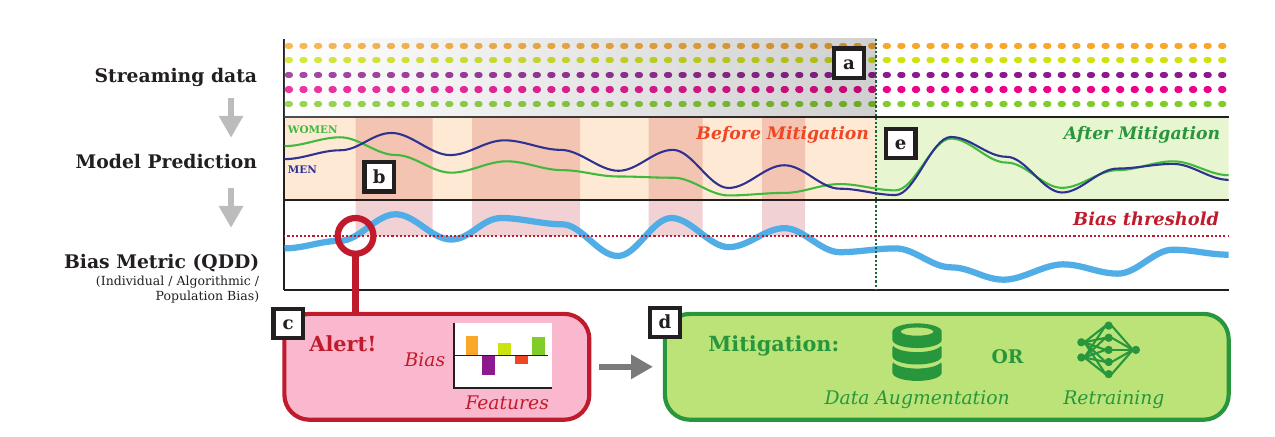}
    \caption{A diagram illustrating how FairCanary monitors the inputs and outputs of a trained model over time, identifies bias, alerts the developer, and assists in mitigation. See \autoref{subsec:overview} for further details.}
    \label{fig:flowchart}
\end{figure*}

\subsection{Shortcomings of Conventional Fairness Metrics}
\label{sec:conventional}

Several conceptual definitions of fairness have been discussed in the literature that, according to~\citet{corbett2018measure}, fall into three general classes: (1) \textit{anti-classification}, where protected features and their proxies are not used to make decisions, (2) \textit{classification parity}, where measures of model predictive performance are equal across protected groups, and (3) \textit{calibration}, where the outcomes, conditional on priors, are independent of protected features. Corbett-Davies and Goel dissect fairness metrics that implement these definitions, claiming that they have ``deep statistical limitations''~\cite{corbett2018measure}, with several metrics at odds with one another. 






\autoref{tab:prosconstable} shows an overview of the terminology and limitations of different classes of fairness metrics in the literature. We refer to the first five frameworks (demographic parity, conditional statistical parity, equalized odds, equal opportunity, and counterfactual fairness) as ``conventional'' fairness metrics because of their prevalence in algorithmic fairness literature~\cite{mehrabi2019survey} and in the industry~\cite{bellamy2018ai}. The last two classes, statistical independence and distributional difference, are relatively niche and new to the discussion.

Conventional fairness metrics have impossibility results~\cite{21fairness}. Prior work~\cite{kleinberg2016inherent,corbett2018measure,miroshnikov2020wasserstein} points out that it is impossible to satisfy both \textit{classification parity} and \textit{calibration} metrics at the same time in general, and therefore context becomes key when picking a metric~\cite{selbst2019fairness,barocas2021designing}.


These statistical limitations extend to group membership limitations. Conventional fairness metrics require groups and subgroups to be discrete variables and cannot work with continuous variables~\cite{ghosh2021characterizing}. Similarly, ``confusion matrix based-metrics''~\cite{21fairness} do not support continuous outputs (which is often the case in problems like regression and recommendation). This limitation necessitates that practitioners choose thresholds for determining if the given fairness metric has been violated, but the process for choosing these thresholds is ad hoc and may lead to wildly different conclusions about the fairness of a model. We show an example of this phenomenon in \autoref{fig:threshold}.


The fairness metric we developed for FairCanary is called Quantile Demographic Drift (QDD). It is a quantile-based optimized version of the Wasserstein-1 distance metric~\cite{villani2009wasserstein}. It falls under the distributional difference family (see \autoref{tab:prosconstable}) and thus lends itself to continuous measurement and explainability. We describe in detail the advantages of QDD over other metrics in \autoref{sec:qdd}.

\section{FairCanary}
\label{sec:overview}

We now describe FairCanary, our system for performing continuous model monitoring. First, we present the context in which FairCanary is designed to operate and describe its operations at a highlevel. Next, we discuss how FairCanary measures bias and introduce our novel Quantile Demographic Disparity (QDD) metric. Finally, we describe how FairCanary provides explanations that attribute observed biases to specific features, and the bias mitigation options provided by FairCanary. 

\subsection{Overview}
\label{subsec:overview}

FairCanary is a system for performing continuous model monitoring. It is designed to be deployed into production environments alongside a trained ML model to help the developers monitor the model's performance over time in terms of traditional and fairness performance metrics. In this paper, our focus will be on the latter, fairness metrics.

The developer of the model must configure FairCanary, a priori, by defining the (intersectional) groups for which unfairness will be monitored, identifying the feature(s) in the dataset that encode group membership, establishing base rate statistics for these groups (\ie as ascertained from the model's underlying training data), and setting thresholds to trigger bias alerts.

\autoref{fig:flowchart} illustrates FairCanary's mode of operation and some of its key capabilities. (\textbf{a}) As new data arrives it is fed into the trained model, which (\textbf{b}) produces predictions that are stored by FairCanary. Over time, FairCanary maintains a record of the predictions for each group at an operator-specified time granularity.

(\textbf{c}) Periodically, FairCanary computes the fairness metric (QDD, see \autoref{sec:qdd}) for the model and alerts the developers if any group performs below the preconfigured threshold. FairCanary provides explanations along with alerts that inform the developer which feature(s) are attributable to the issue (see \autoref{sec:attribution}). (\textbf{d}) Subsequently, the developer may mitigate the emergent unfairness using tools provided by FairCanary (see \autoref{sec:mitigation}), which (\textbf{e}) should return the model to a state where predictions are fair across groups.


\subsection{Quantile Demographic Disparity}
\label{sec:qdd}

In this section, we describe a new metric to measure bias in the predictions of a ML model, at both the group and individual level. The prediction tasks covered by our metric include any single dimensional output, such as regression output, or the output of any particular class in a multi-class classification model.

Our metric, Quantile Demographc Disparity (QDD), falls within the distributional difference family of fairness metrics (see \autoref{tab:prosconstable}). We argue that there are two reasons for assessing the fairness of an ML model by comparing its prediction distributions over the groups of interest, versus focusing on post-threshold outcomes. The \textit{first} reason is to ensure that we measure bias across the whole spectrum of classified individuals, as opposed to focusing solely on the individuals that are above the threshold of selection, or on group-level approximations. \textit{Second}, as groups of interest get smaller, they reveal more information about intra-group disparities that would have otherwise been lost due to aggregation~\cite{ghosh2021characterizing}, all the way down to groups of one, \ie individuals. This helps remove aggregation bias from the bias measurement itself.

\subsubsection{\textbf{Desired Properties of a Bias Metric}}
\label{sec:properties}

We now discuss desirable properties of a distributional fairness metric that fit our stated objectives:
\begin{enumerate}
    \item The metric should be in the units of the model's prediction scores. The utility of this is especially evident when dealing with continuous output models. This is desirable because it provides insight into the extent of the problem, before human intervention is applied, such as deciding and applying a threshold.
    \item The metric should take the value zero only if the prediction distributions being compared are exactly the same. The benefit of this is that, when taken along with the first property, it gives the ML practitioner a mental scale to understand the extent of the bias.
    \item The metric should be continuous with respect to changes in the geometry of the distribution~\cite{miroshnikov2020wasserstein}. This ensures that any distributional change is captured.
    \item The metric should be non-invariant with respect to monotone transformations of the distributions~\cite{miroshnikov2020wasserstein}. For example, given two samples of points $S_1$ and $S_2$, if we multiply the value of each point in the samples by a constant $k$, the distance between the modified samples should now depend on $k$. Jensen-Shannon Divergence (JSD)~\cite{lin1991divergence}, for example, does not satisfy this property.
    \item The metric should be bias-transforming as described in~\cite{wachter2020bias}, \ie the metric should not be satisfied by a model that preserves the biases present in the data.
\end{enumerate}
QDD satisfies all of these properties when the number of bins is equal to the number of samples.The choice of number of bins can be adjusted to satisfy these properties.

\subsubsection{\textbf{Formalization}}\label{sec:theory}

We now describe our QDD metric, which is a function of the quantile bin that a prediction event lies in. QDD is a novel formulation of the Wasserstein-1 distance metric~\cite{villani2009wasserstein}, and thus it is designed to work for continuous outputs and can be customized to provide sliced views down to the individual-level.

For two groups $G_{1}$ and $G_{2}$, let the two distributional samples of model scores be $S_{1}$ and $S_{2}$. We divide the samples into $B$ bins of equal size $N_{1}$ and $N_{2}$, respectively. This is equivalent to segmenting by quantiles. For example, if there are 10 bins, we are essentially bucketing individuals between the \nth{0}--\nth{10} percentile, \nth{10}--\nth{20} percentile, and so on. 

We define QDD for bin $b$ as
\begin{equation}
QDD_{b} = \mathbb{E}_{G_{1, b}}[S_{1}] - \mathbb{E}_{G_{2, b}}[S_{2}].
\end{equation}
This can be approximated as
\begin{equation}
QDD_{b} = \frac{1}{N_{1}}  \sum\limits_{n=1}^{N_{1}} S_{1, n} -  \frac{1}{N_{2}}  \sum\limits_{n=1}^{N_{2}} S_{2, n}.
\end{equation}
The QDD, when conditioned on certain attributes $C$, becomes the Conditional Quantile Demographic Disparity.

To demonstrate the flexibility of QDD, we demonstrate how it can be used to measure three different conceptualizations of bias.

\noindentparagraph{(1) Intra-Group Bias} is defined as the maximum QDD across the $b$ bins of a given group of individuals. This quantity is useful to combat aggregation bias within groups.

\noindentparagraph{(2) Disparity with Base Rate} is defined as the difference between the QDD calculated over the production data and the QDD of the training data. This quantity is most relevant when the training data is representative of the population the model is expected to encounter during deployment. 

\noindentparagraph{(3) Individual Fairness via Alignment.}

QDD is defined between two groups over a given number of bins, which determines the resolution of the metric. If the number of bins is equal to the number of instances in the sample, QDD becomes a comparison between individuals at the same rank or percentile. This is equivalent to the concept of \textit{alignment} proposed by~\citet{shanbhag2021unified}.

Computing QDD over individual instances gives us a clean way to obtain individual fairness insights, with the counterfactual example being the same ranked counterpart in the opposite group. This method does not require us to compute complex counterfactuals, which could have their own biases and errors~\cite{kilbertus2020sensitivity}. The principle we use to justify this insight is as follows: if there is no bias between two groups, and we have a large enough sample of both, then the distance between individuals of the same rank in the prediction space should be zero.


\subsection{Explanation}
\label{sec:attribution}

Explainability of ML systems that are deployed in production is a very important part of the practice of responsible AI~\cite{markus2021role,bhatt2020explainable}. This especially applies to models that are contributing to decisions that can impact peoples' lives. Such decisions cannot be inscrutable, and thus the internal workings employed by the ML models must be human-verified to be logical and normatively justifiable.

FairCanary incorporates two state-of-the-art methods for explaining the output of predictions in terms of specific features: Shapley value-based methods~\cite{shapnips} and (if the model being monitored is differentiable) Integrated Gradients (IG)~\cite{sundararajan2017axiomatic}. We adopted these methods because they satisfy the desirable axiom of \textit{efficiency}~\cite{shapnips}, which helps provide a precise accounting of bias.\footnote{Although there are other explanation methods that satisfy efficiency~\cite{shrikumar2017learning, binder2016layer} we do not explore them in this work.} 



Just like explanations for individual predictions, we argue that it is vital to be able to explain measures of fairness or bias, so that the features that are responsible for the bias can be identified. FairCanary incorporates a novel method for explaining the feature importance contributing to QDD that we call Local Quantile Demographic Disparity attribution. 

The Local QDD attribution for feature $f$, for prediction sample $S_{1}$ over $S_{2}$ in bin $b$, \(QDDA_{b, A, f}\) is a measure of the change in QDD in bin $b$ that can be attributed to (\aka explained by) feature $f$ using attribution method $A$ that satisfies the efficiency axiom. $r$ denotes the reference and $t$ denotes the target distribution. We define
\begin{equation}
\text{QDDA}_{b, A, f} = \frac{1}{N_{t}}  \sum\limits_{n=1}^{N_{t}} \text{attr}_{n, A,S_{1},f} -  \frac{1}{N_{r}} \sum\limits_{n=1}^{N_{r}} \text{attr}_{n, A, S_{2},f}
\end{equation}
where \(\text{attr}_{n, A, S_{i}, f}\) refers to the attribution of the $n^{\text{th}}$ data point to feature $f$ for a prediction from bin $b$ of distribution $S_{i}$ using attribution method $A$. Given that the attribution method $A$ satisfies the efficiency axiom,  $\text{QDD}_{b} = \sum \limits_{f=1}^{F} \text{QDDA}_{b, A, f}$.
 
\textit{Proof:} 
Since the attribution method $A$ satisfies efficiency, for each instance in the sample $S_{1}$ and $S_{2}$,
$\sum\limits_{f=1}^{F} \text{attributions}_{f} = \text{prediction} - \text{baseline prediction}$.

For the same baseline, 
\begin{equation*}
\begin{split}
\frac{1}{N_{t}}\sum\limits_{n=1}^{N_{t}} \sum\limits_{f=1}^{F}  \text{attr}_{n, A,S_{1}, f} -  \frac{1}{N_{r}}\sum\limits_{n=1}^{N_{r}} \sum\limits_{f=1}^{F}  \text{attr}_{n, A,S_{2}, f} = \\
\frac{1}{N_{1}}  \sum\limits_{n=1}^{N_{2}} S_{1, n} -  \frac{1}{N_{2}}  \sum\limits_{n=1}^{N_{2}} S_{2, n}
\end{split}
\end{equation*}
\begin{equation*}
\therefore \text{QDD}_{b} = \sum \limits_{f=1}^{F} \text{QDDA}_{b, A, f}.
\end{equation*}
Explaining bias in this manner enables a single attribution to be used for multiple explanations across groups. In contrast, Shapley values over a particular metric must be re-calculated for every grouping. Our explanation technique therefore is much more computationally efficient than previous techniques~\cite{miroshnikov2020wasserstein} since it requires the calculation of attributions only once. To elaborate, Shapley values without approximation are exponential in the number of model features. While there exist approximation techniques, the complexity is worse than linear time. Hence, Wasserstein Shapley computation (n points times d features) is worse than calculating Shapley values for n points separately for d features. Additionally, we can re-use the Shapley values computed for a data point when calculating QDD between any combination of protected features, whereas the whole computation needs to be repeated for each combination in the case of Wasserstein Shapley.





\subsection{Mitigation}
\label{sec:mitigation}

Mitigation is a key outcome of monitoring bias, enabling corrective action to be taken. FairCanary provides an option for developers to automatically mitigate bias revealed by our QDD metric using a quantile norming approach. In essence, this approach replaces the score of the disadvantaged group with the score of the corresponding rank in the advantaged group, similar to the mitigations proposed in~\cite{jiang2020wasserstein, nandy2021achieving}. The justification for quantile norming is that if (1) bias is known to exist, (2) bias is the only rational explanation for disparity, and (3) bias is assumed to be equal within the disadvantaged group, then normalizing across ranks is normatively justifiable.

In essence, quantile norming is a post-processing mitigation. The advantages of post-processing mitigations as opposed to pre-training debiasing are discussed by~\citet{geyik2019fairness}. Additionally, quantile norming is a relatively computationally inexpensive approach to bias elimination.

We note that all bias mitigation approaches, including quantile norming, should only be adopted in practice after conducting a thorough examining of their consequences on the outputs of a model.~\citet{corbett2017algorithmic} demonstrate several cases where mitigation may cause additional harm to individuals or to particular groups. Developers that adopt FairCanary are under no obligation to use quantile norming for mitigation, and are free to adopt other, perhaps more thorough and computationally expensive, approaches (\eg model retraining~\cite{schelter2018challenges}, data preprocessing~\cite{kamiran2012data}, etc.) that better suit their needs.


 

\section{Case Study}
\label{sec:casestudy}


\begin{table}[t]
    \centering
    \resizebox{\linewidth}{!}{%
    \begin{tabular}{lll}
    \toprule
    \textbf{Feature} & \textbf{Values} & \textbf{Distribution} \\
    \midrule
    Location & \{`Springfield', `Centerville'\} & 70:30 \\
    Education & \{`GRAD', `POST\_GRAD'\} & 80:20 \\
    Engineer Type & \{`Software', `Hardware'\} & 85:15 \\
    Experience (Years) & $(0, 50)$ & Normal Distribution\\
    Relevant Experience (Years) & $(0, 50)$ & Normal Distribution \\
    Gender & \{`MAN', `WOMAN'\} & 50:50 \\
    \bottomrule
    \end{tabular}
    }
    \caption{Features, values, and their distributions used in our synthetic case study. Note that the gender feature is only used for measuring and mitigating bias, it is not used for model training or prediction.}
    \label{tab:features}
    \vspace{-0.5cm}
\end{table}

\begin{figure*}[t]
    \begin{subfigure}[t]{0.3\textwidth}
        \centering
        \includegraphics[width=\columnwidth,keepaspectratio]{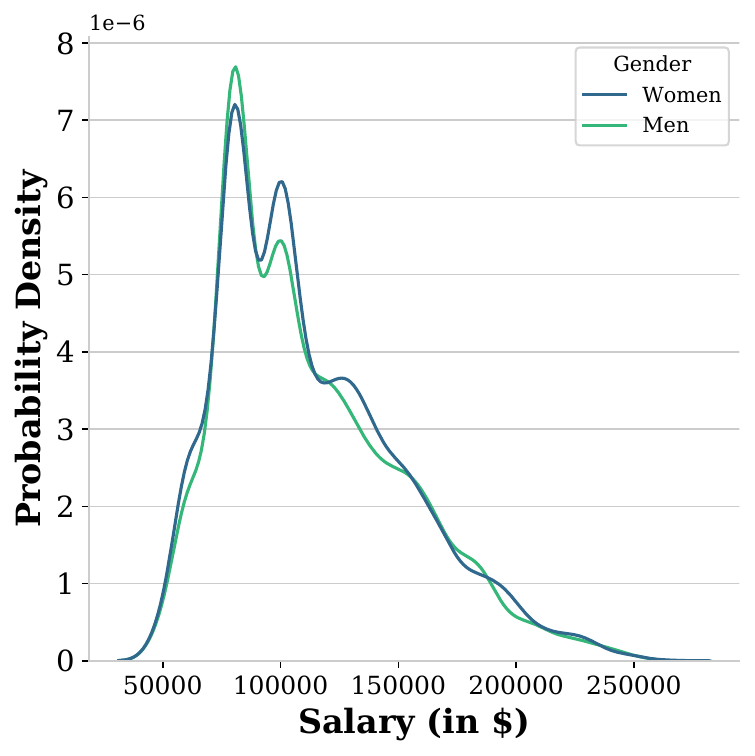}
        \caption{Prediction distribution on Day One}\label{fig:preday1}
    \end{subfigure}
    \begin{subfigure}[t]{0.3\textwidth}
        \centering
        \includegraphics[width=\columnwidth,keepaspectratio]{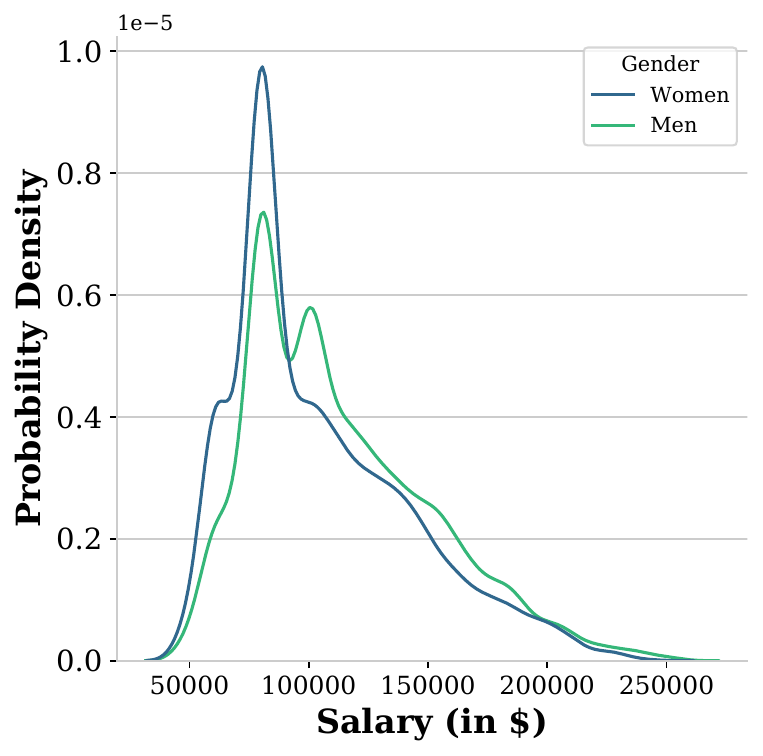}
        \caption{Prediction distribution on Day Two}\label{fig:preday2}
    \end{subfigure}
    \begin{subfigure}[t]{0.3\textwidth}
        \centering
        \includegraphics[width=\columnwidth,keepaspectratio]{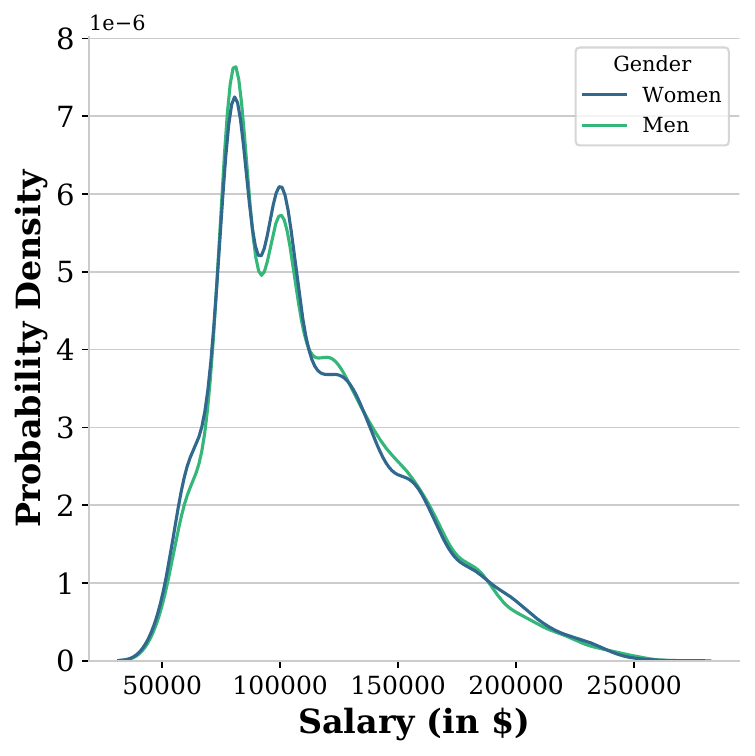}
        \caption{Prediction distribution on Day Three}\label{fig:preday3}
    \end{subfigure}
    \begin{subfigure}[t]{0.9\textwidth}
    \centering
        \includegraphics[width=\textwidth,keepaspectratio]{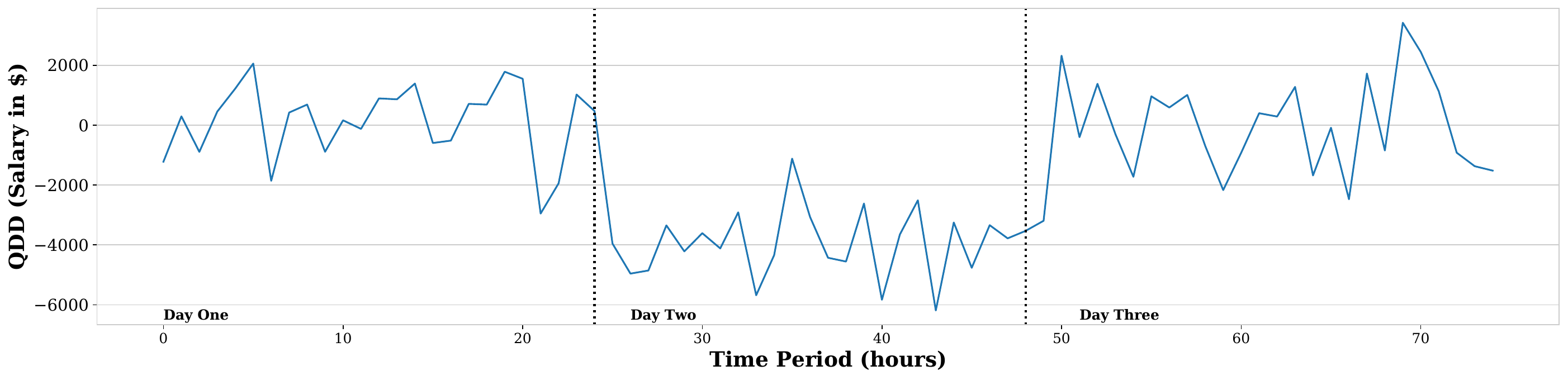}
        \caption{Continuous plot of the QDD metric over time. There is a clear dip on the second day.}\label{fig:qddcasestudy}
    \end{subfigure}
    \begin{subfigure}[t]{0.3\textwidth}
        \centering
        \includegraphics[width=\columnwidth,keepaspectratio]{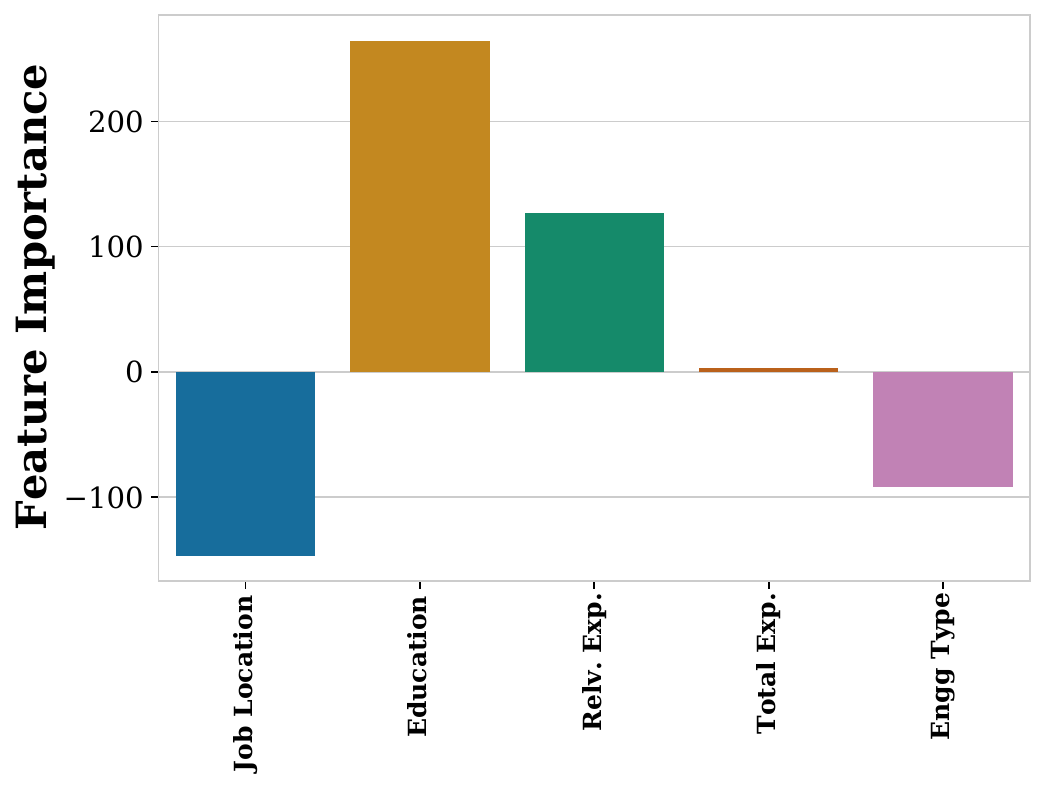}
        \caption{QDD Explanations for Day One}\label{fig:expday1}
    \end{subfigure}
    \begin{subfigure}[t]{0.3\textwidth}
        \centering
        \includegraphics[width=\columnwidth,keepaspectratio]{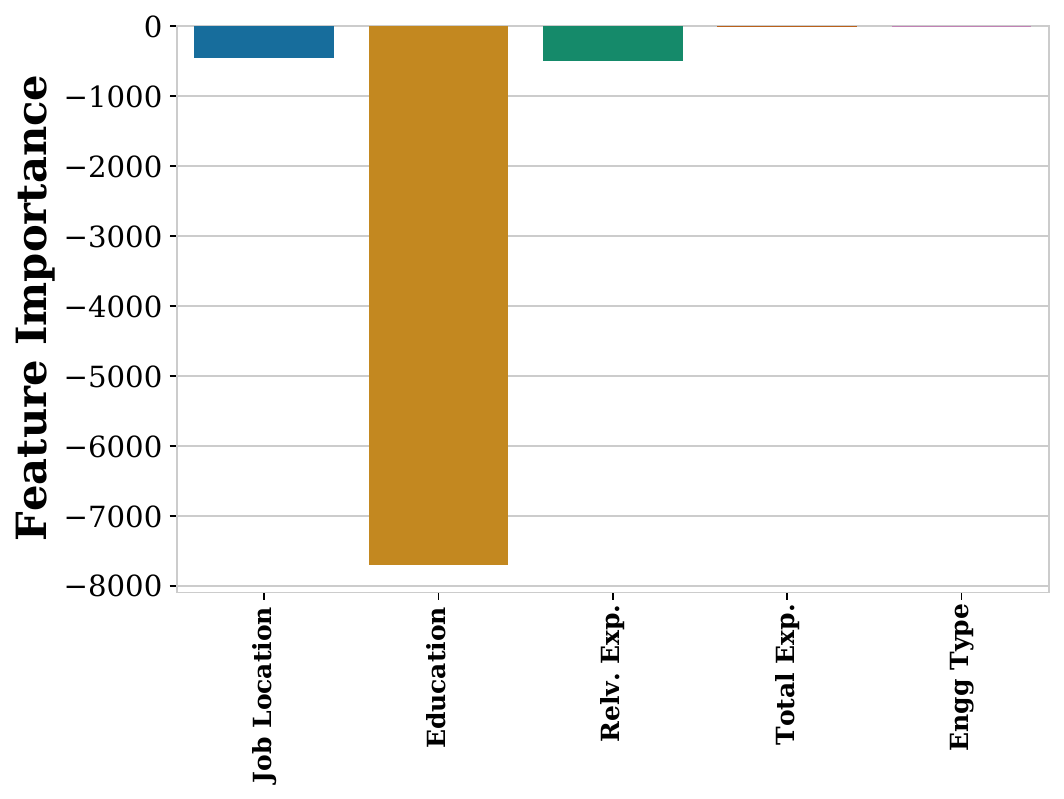}
        \caption{QDD Explanations for Day Two}\label{fig:expday2}
    \end{subfigure}
    \begin{subfigure}[t]{0.3\textwidth}
        \centering
        \includegraphics[width=\columnwidth,keepaspectratio]{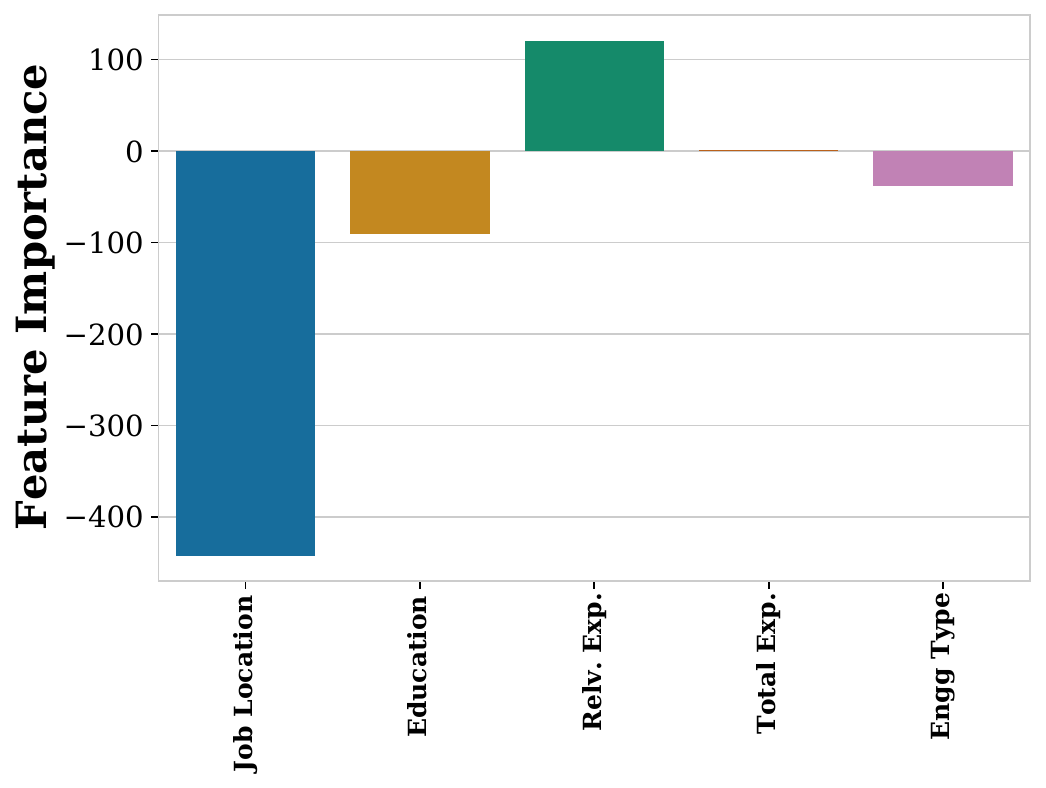}
        \caption{QDD Explanations for Day Three}\label{fig:expday3}
    \end{subfigure}
    \caption{Plots for our case study showing how FairCanary would detect and explain the bias against women on Day Two on a continuously running salary prediction model. The explanations for Day Two clearly indicate Education as the feature responsible for the bias, which enables the practitioner to correct the data integrity issue and fix the biased predictions.}
    \label{fig:casestudy}
\end{figure*}\textbf{}




In this section, we present an example of FairCanary in action via a case study on a synthetic dataset. This allows us to inject controlled drifts into the data stream to demonstrate how FairCanary, via QDD, can detect and explain the resulting bias. Additionally, we present comparisons to conventional fairness metrics.

\subsection{Scenario}

In this case study, we posit a scenario where a developer has trained a model to predict the starting salary of job seekers based on relevant features of their resume, such as education level and years of experience (see \autoref{tab:features}). Note that the output of this model is continuous. Additionally, the developer designed the model to be fair with respect to the binary gender of job seekers, \ie the distribution of salaries predicted for men and women should be nearly identical. We assume that the model was audited and found to be fair relative to the data that was available at training time.

Let us assume that the model has learned the following relationship to predict an individual's salary from the features in  \autoref{tab:features}:
\begin{equation*}
\begin{split}
\text{Salary} = 50,000 + (20,000 \times \text{location}) + (20,000\\ \times  \text{education}) + (5,000 \times \text{relevant\_experience})\\ + (100 \times \text{experience}) +(10,000 \times \text{engineer\_type})
\end{split}
\end{equation*}

In our scenario, the developer deploys this model into production along with FairCanary to continuously monitor its output. We generate 20,000 synthetic job seekers' data per day for three days that are fed into the model, using feature values drawn from the distributions given in \autoref{tab:features} (with the added constraint that experience $\geq$ relevant experience). On Day One and Day Three we generate all of the candidate data correctly, but crucially, on Day Two, we simulate a data engineering bug that erroneously labels all women as `GRAD' instead of `POST\_GRAD' regardless of their true educational attainment. This reduces the estimated salary for all women post-graduates by \$20,000 on Day Two.

We argue that the scenario we have outlined here is realistic. ML-based resume screening and analysis tools are widely available, and given that they gate access to employment opportunities, it is crucial that these systems be fair~\cite{bogen2018help,raghavan-2020-fat}. The bug we intentionally simulate on Day Two could easily occur in practice, \eg due to the temporary malfunctioning of a resume parser that prepares data for the salary prediction model.

\subsection{Analysis}

\autoref{fig:casestudy} shows how FairCanary would detect and explain the fairness problem that occurs on Day Two. The model outputs on Day One show that the prediction distributions for men and women are mostly aligned (\autoref{fig:preday1}), thereby being fair. On the second day (\autoref{fig:preday2}), due to the data integrity error discussed above, the prediction distributions differ. When we examine the running plot for QDD\footnote{For simplicity, we set the number of quantile bins as 1 for the case study. Thus, the explanations are for the entire distribution and not any particular quantile bin.} (\autoref{fig:qddcasestudy}), we notice a sharp dip on Day Two---QDD goes from an average value of \$156 on Day One to -\$8677 on Day Two---indicating a bias against women.\footnote{Recall in \autoref{sec:properties} we say that one useful feature of QDD is that the metric has the same units as the predicted output. Having the QDD value in dollars clearly helps users to understand the extent of bias and thereby aids usability.} Note that the absolute value of QDD goes up, indicating an increase in bias, and would trigger the alarm system like in \autoref{fig:flowchart}. Similarly, the feature explanations (generated here using Integrated Gradients) go from being distributed among the different features on Day One (\autoref{fig:expday1}) to assigning the majority of blame to the education feature on Day Two (\autoref{fig:expday2}).

FairCanary would alert the model developer of the problem on Day Two, and its explanations could help the developer perform root-cause analysis of the bias issue. Based on this information, the developer could then identify and correct the underlying data engineering bug. Once corrected, we observe that the model's predictions are again aligned for men and women on Day Three (\autoref{fig:preday3}), and the QDD values have returned to their expected range (\autoref{fig:qddcasestudy}).

\begin{table}[t]
\centering
\begin{tabular}{lllll}
\toprule
\multicolumn{1}{c}{\textbf{}} & \multicolumn{2}{c}{\textbf{Day One}} & \multicolumn{2}{c}{\textbf{Day Two}} \\
\cmidrule(lr){2-3} \cmidrule(lr){4-5}
\multicolumn{1}{l}{\textbf{Threshold}} & \textbf{SPD} & \textbf{DI} & \textbf{SPD} & \textbf{DI} \\
\midrule
\$50000  & \cellcolor{green!25}0.00009      & \cellcolor{green!25}1.00009     & \cellcolor{green!25}-0.00556     & \cellcolor{green!25}0.99439    \\
\$100000 & \cellcolor{green!25}0.00911      & \cellcolor{green!25}1.01749     & \cellcolor{green!25}-0.08290     & \cellcolor{green!25}0.84569    \\
\$200000 & \cellcolor{green!25}0.00088      & \cellcolor{green!25}1.02876     & \cellcolor{green!25}-0.01049     & \cellcolor{red!25}0.65544 \\
\bottomrule
\end{tabular}
\caption{The performance of two conventional fairness metrics, Statistical Parity Difference (SPD) and Disparate Impact (DI), against different salary thresholds for the case study. The predictions on Day One were fair, while they were unfair to women on Day Two. Only one metric catches the bias, and only at one threshold (highlighted in red).}
\label{tab:comparison}
\vspace{-0.5cm}
\end{table}

To further illustrate the utility of QDD we compare it with two conventional fairness metrics---Statistical Parity Difference (SPD)\footnote{Statistical or Demographic Parity Difference is the difference in the positive outcome rate between the privileged and unprivileged group. SPD = Pr$(\hat{y} = 1 | p = 1)$ - Pr$(\hat{y} = 1 | p = 0)$.}, and Disparate Impact (DI)\footnote{Disparate Impact is the ratio of the passing rate of the the privileged and unprivileged group. DI = $\frac{Pr(\hat{y} = 1 | p = 1)} {Pr(\hat{y} = 1 | p = 0)}$.}---to see if a monitoring system using these metrics would have caught the bias against women on the second day.

\autoref{tab:comparison} shows the values of the two conventional metrics for different salary thresholds (\ie for the positive outcome) on Day One and Day Two. We configure the alert threshold for both metrics\footnote{For Statistical Parity Difference, since there is no conventionally accepted value, we set the threshold to 20\% to be consistent with Disparate Impact.} in accordance with the US UGESP 4/5$^{\text{ths}}$ rule~\cite{ugesp} that is commonly used in disparate impact analysis~\cite{wilson2021building}. Alarmingly, we observe that, as configured here, SPD would not catch the bias on the second day at all, and DI would only catch it at one threshold level.

\section{Discussion}
\label{sec:discussion}

In this work we present a novel metric called QDD that improves on conventional fairness metrics by not requiring prediction labels or threshold values (\autoref{sec:qdd}). We utilize this metric in FairCanary, a system for performing continuous monitoring of deployed ML models. FairCanary includes all of the typical capabilities of ML monitoring systems~\cite{neptune}: it records inputs to and outputs from the model over time, calculates traditional measures of model performance (\eg accuracy), allows operators to set configurable alerts if model performance changes dramatically, and calculates explanations for individual predictions using existing techniques~\cite{shapnips,sundararajan2017axiomatic}.

Additionally, FairCanary is able to provide explanations for QDD by reusing the explanations for individual predictions, which is (1) a capability not offered by conventional fairness metrics and (2) less computationally demanding than similar approaches from prior work~\cite{miroshnikov2020wasserstein} (\autoref{sec:attribution}).

Through examples (\autoref{fig:threshold}) and a synthetic case study (\autoref{sec:casestudy}), we demonstrate the functionality of FairCanary and the useful properties afforded by our QDD metric. We publicly release the code \footnote{\url{https://github.com/fiddler-labs/faircanary}} used to generate the plots in our case study.

\subsection{Limitations and Future Work}

While threshold independence is one of the strengths of QDD, it is also a potential weakness: without ground truth labels, calculated disparities are, at the end of the day, best-case approximations of the discrimination that actually takes place in society. We therefore do not advocate for the elimination of conventional fairness metrics that require ground truth labels and thresholds, but instead propose using them in conjunction with QDD to obtain a fuller picture of real life harms in a context dependent manner~\cite{aequitas}.

We used Integrated Gradients as the explanation method for our case study. However, the choice of explanation method is potentially important, as recent research~\cite{hooker2019benchmark,pmlr-v119-kumar20e} shows that different explanation methods often do not produce the same results, and ensembling them is superior than using any one of them in isolation.

Finally, FairCanary/QDD is not completely automated: there are still manual parameters that need to be set, like number of bins and alert sensitivity. Providing FairCanary users with guidance on how to tune the system for their use case and context will be crucial for real use cases. Additionally, all fairness monitoring systems should consider providing actionable recourse tips~\cite{karimi2020survey} through explanations to end-users via a carefully designed, accessible interface. 

\subsection{Broader Impact}

Regardless of whether ML models are regulated to mandate audits and continuous monitoring, we argue that ML practitioners have a professional and moral obligation to ensure that the systems they deploy do not misbehave. Given that issues like drift are known to occur, and that these issues may cause unfairness and bias, we argue that monitoring systems should become a standard component of most, if not all, deployed ML-based systems.

We hope that FairCanary (or other monitoring systems that incorporate its capabilities) will equip companies and institutions with improved tools to monitor, understand, and mitigate problems in their deployed ML systems, in real time. In turn, we hope that these capabilities will bring more equity and justice to the individual stakeholders impacted by deployed models.

\begin{acks}
The authors would like to thank Hima Lakkaraju, Josh Rubin, and Shobhit Verma at Fiddler Inc. for their constructive suggestions during the ideation and improvement of this work, specifically for the synthetic experiment. Additionally, the authors  thank Jane L. Adams, Jeffrey Gleason, and Michael Davinroy for their comments on the draft of the paper, and Jane L. Adams for her assistance in creating \autoref{fig:flowchart}.
\end{acks}

\balance
\bibliographystyle{ACM-Reference-Format}
\bibliography{ref}

\end{document}